# Adversarial Semi-supervised Learning for Corporate Credit Ratings

Bojing Feng[1][2], Wenfang Xue[1]*

[1] Center for Research on Intelligent Perception and Computing, National Laboratory of Pattern Recognition, Institute of Automation, Chinese Academy of Science. Beijing China
[2] School of Artificial Intelligence, University of Chinese Academic of Science

* Corresponding author. Tel.: +8613683573380; email: wenfang.xue@ia.ac.cn

**Abstract:** Corporate credit rating is an analysis of credit risks within a corporation, which plays a vital role during the management of financial risk. Traditionally, the rating assessment process based on the historical profile of corporation is usually expensive and complicated, which often takes months. Therefore, most of the corporations, due to the lack in money and time, can't get their own credit level. However, we believe that although these corporations haven't their credit rating levels (unlabeled data), this big data contains useful knowledge to improve credit system. In this work, its major challenge lies in how to effectively learn the knowledge from unlabeled data and help improve the performance of the credit rating system. Specifically, we consider the problem of adversarial semi-supervised learning (ASSL) for corporate credit rating which has been rarely researched before. A novel framework adversarial semi-supervised learning for corporate credit rating (ASSL4CCR) which includes two phases is proposed to address these problems. In the first phase, we train a normal rating system via a machine-learning algorithm to give unlabeled data pseudo rating level. Then in the second phase, adversarial semi-supervised learning is applied uniting labeled data and pseudo-labeled data to build the final model. To demonstrate the effectiveness of the proposed ASSL4CCR, we conduct extensive experiments on the Chinese public-listed corporate rating dataset, which proves that ASSL4CCR outperforms the state-of-the-art methods consistently.

## 1. Introduction

Nowadays, corporate credit rating is a crucial technology to help the broad masses of the people and financial institutions such as Standard & Poor's (S&P's), Moody's and Fitch Ratings to mitigate credit risk. Credit rating not only is an indication of the level of the risk but also represents the probability that the corporation pays its financial obligations on time [1]. The credit rating method undergoing reforms has made significant progress and achieved remarkable results [2]. Traditionally, rating agencies perform an assessment including qualitative and quantitative with many experts involved to analyze all kinds of variables to get the credit level of a company, which is usually very expensive and complicated. Therefore, it is of great importance to build a credit rating system that can model the profile of the corporation and predict the credit level.

Some credit risk models have been proposed since the middle of the twentieth century [3]. Traditionally,

most of models are data analysis and statistical technologies. Researchers apply these methods into credit risk problem such as logistic regression, Markov models. Nowadays, machine learning and deep learning models have shown their power in most of fields especially the financial field. These methods are from classical machine learning algorithms such as random forest to neural networks. Now in deep learning, it begins with Muti-Layer Perception (MLP), going through Convolution Neural Networks (CNN) to Long-Short-Term-Memory (LSTM). Recently, the self-attention mechanism architectonics such as transformer is also applied in this field and graph-based methods including Graph Neural Networks (GNNs) have been tried in this field. All kinds of models from multi-views analyze and model this problem to improve performance continuously and to be the stat-of-the-art-model.

Although neural networks and deep learning methods can efficiently fit high-dimensional and non-linear data. However, there is a premise that all these models need thousands of data to train models and optimize the parameters. In fact, due to the nature of financial peculiarly in corporate credit rating problem, only few corporations have their own credit level given by credit agency. In terms of small and medium-sized enterprises (SMEs), on the one hand, the scale of the enterprise is small and various financial information is not perfect. On the other hand, they can hardly afford the time waste and money cost during the assessment of credit rating within rating experts. Isn't this unlabeled data worthless? Of course, our answer is No. We believe there is a lot of knowledge under unlabeled data. The major challenge of this work lies in how to effectively mine the knowledge from unlabeled data to improve the performance of credit rating model.

To overcome the problems described above and fully take advantage of neural networks, we proposed a novel end-to-end framework inspired by Generative Adversarial Networks (GAN) and semi-supervised learning, adversarial semi-supervised learning for corporate credit rating, ASSL4CCR for brevity. Different from previous credit rating methods, we firstly combine the adversarial networks and semi-supervised learning applying in the field of corporate credit rating.

The ASSL4CCR framework is composed of two phases. In the first phase, the pseudo label for unlabeled data is given by training a normal credit rating model such as mlp, gbdt, xgboost in order to learn the knowledge from lots of unlabeled dataset. In the beginning of this work, we find that only directly applying semi-supervised learning by pseudo-labeled dataset suffer from the misalignment of representations between supervised task and semi-supervised task. This may weaken the representation generalization of semi-supervised corporate credit rating model. To tackle these problems, we make two improvements during the second phase. Firstly, we build an encoder function at the beginning of the second phase to map the labeled data and unlabeled into same vector space for mitigating the misalignment between supervised and semi-supervised task. Moreover, inspired by the effect of adversarial learning, we exploit its application to couple the adversarial learning into supervised and semi-supervised learning.

For clearly illustrating our main idea, we make an example as an analogy here. A student (credit rating model) wants to learn. His teachers will give him partly standard answers (labeled data). However, most of non-standard answers are from his classmates (pseudo-labeled data). He wants to learn from both standard answer and non-standard in order to make a perfect test. Fortunately, he has the ability of distinguishing where the answer from teachers or classmates (adversarial learning). For the answer from teachers, he believes without doubt. While others, he is supposed to believe them partly. Under this strategy, he can make the best test as far as possible.

To sum up, the main contributions of this work are summarized as follows:
- We propose an Adversarial Semi-Supervised Learning for Corporate Credit Rating (ASSL4CCR) framework to learn the useful knowledge from unlabeled data, which firstly try to tightly couple adversarial learning and semi-supervised learning in the field of corporate credit rating.
- During the studying, we find that only semi-supervised learning by pseudo-labeled dataset could

incur the misalignment of representations between supervised and semi-supervised task. Therefore, two novel strategies, encoder function and adversarial learning, are designed to couple adversarial learning into corporate credit rating, which further boosts the power of the representation and generalization.
- Comprehensive experiments on the Chinese public-listed corporate rating dataset demonstrate that our framework can exactly learn the knowledge from unlabeled data to improve the performance of credit rating models experimentally.

This paper is organized as follows. Related works about corporate credit rating and adversarial semi-supervised learning are introduced in Section 2. Section 3 will present the proposed framework ASSL4CCR. The experiment results of ASSL4CCR on real-word data will be shown in Section 4. Finally, the conclusion is in Section 5.

## 2. The Related Works

In this section, we firstly review some related works on credit risk, including statistical and machine learning models, deep learning models and graph-based models. Then we introduce some works about adversarial learning and semi-supervised learning.

### 2.1. Corporate Credit Risk

#### 2.1.1. Statistical and Machine Learning Models

Traditionally, researchers apply some statistical and machine learning models such as probit regression in bank credit rating. Recently, a rating model based on Student's-t Hidden Markov Models (SHMMs) [4] was used to investigate the data of firm. There are some researches applying adaptive learning networks (ALN) and support vector machine to perform credit rating respectively. Besides, classification and regression tree technique (CART) is also used to predict consumer credit risk.

#### 2.1.2. Deep Learning Models

Nowadays, deep learning technique has shown its power in various fields. In the work [5], Boltzmann machine was used for credit scoring. As convolution neural networks (CNN) achieve great success in computer vision (CV), CNN based models [6]-[7] were also applied in financial prediction. What's more, the work [8] explored the effect of image encoding for CNN in finance. In terms of the sequence model, long-short term memory (LSTM) based model [9] as a deep learning technique was for time series corporate credit score prediction. Besides, after Transformer [10] succeed in the natural language processing (NLP), it also was applied in corporate credit rating integrated with gated recurrent unit (GRU) [11].

#### 2.1.3. Graph-based Models

With the advance of GNN and its expression, graph-based methods attract the interest of many researchers. In the work [12], the graph-structured loan behavior data was used by attention network. Deeptrax [13] was proposed in order to learn the embedding from financial transactions. Cheng et al. [14] proposed HGAR to learn the embedding of guarantee networks. Recently, works [15]-[16] combined spatio-temporal information for financial risk.

### 2.2. Adversarial and Semi-supervised Learning

GAN was firstly proposed in 2014. After that, infoGAN [17], styleGAN [18] and so on came along successively. The core idea of GAN is adversarial learning. Semi-supervised learning is a learning algorithm which learn from both of labeled data and unlabeled data, usually mostly unlabeled. It has been widely applied in the CV and NLP. The work [19] most related to ours leveraged adversarial self-supervised learning for 3D action recognition. In this work, we explore how to extract knowledge from unlabeled data via adversarial semi-supervised learning.

## 3. The Proposed Framework

In this section, we introduce the proposed method ASSL4CCR which combines adversarial learning and semi-supervised learning for corporate credit rating problem. We formulate the problem firstly, then give an overview of ASSL4CCR. Finally, two phases of the proposed model are illustrated.

### 3.1. Problem Formulation

In credit rating, let $C$ to be the training set which includes $C_L = \{c_1, c_2, \ldots, c_L\}$ labeled dataset an $C_U = \{c_1, c_2, \ldots, c_U\}$ unlabeled dataset. $U$ is usually much bigger than $L$. The training set $C$ consisting of all unique corporations. Every corporation $c$ represents its profile data. In terms of $C_L$, every corporation $c$ has a corresponding label which represent its credit level. Let $Y = \{y_1, y_2, \ldots, y_m\}$ denote the rating level set and $m$ represents the size of unique label. For the unlabeled dataset $C_U$, there is no corresponding label. The goal of credit rating model is to predict the credit level according the profile data of corporation. The corporate credit rating model will output the probability of all credit levels $\hat{y}$ for each corporation. The label with max probability will be the final credit level.

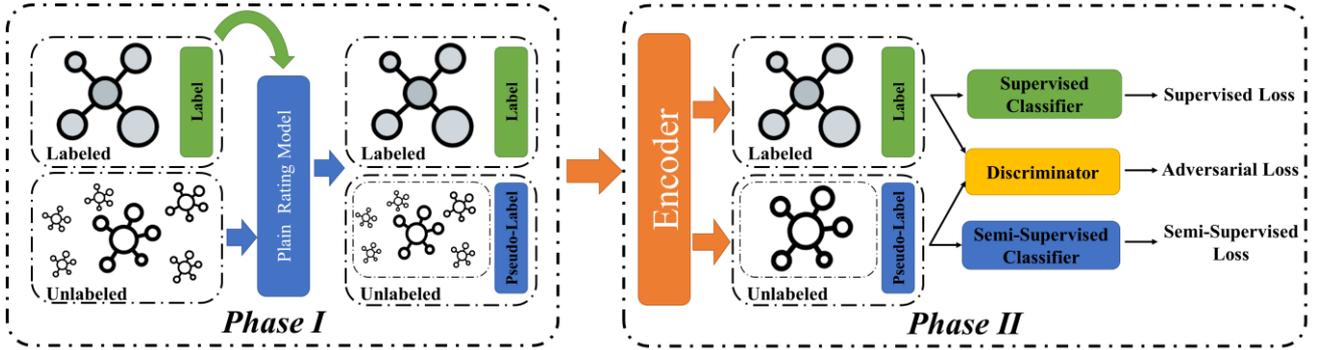

Figure 1: The architecture of our proposed framework ASSL4CCR with two phases.

### 3.2. Model Overview

Figure 1 illustrates our proposed framework ASSL4CCR which includes two phases. During the phase Ⅰ, a plain credit model is trained only by labeled dataset. After that, the pseudo label will be gotten by feeding the unlabeled data into trained plain model. In this way, all the dataset including labeled and unlabeled data could get corresponding labels. Begin with phase Ⅱ, an encoder module is applied to map the data into same space for the misalignment. Both supervised and semi-supervised classifiers perform rating task with producing supervised and semi-supervised loss. In spired by the GAN, the discriminator will tell the model whether the sample is from labeled dataset or unlabeled dataset. The final loss is composed of these three types: supervised loss, semi-supervised loss and adversarial loss.

### 3.3. The Phase Ⅰ

During Phase I, a plain rating model (PRM) is trained only by using labeled data. In theory, any corporate credit rating model can be used within our PRM, such as logistic regression, MLP, GBDT or xgboost. When selecting a machine learning model as PRM, we need to take balanced between model complexity and model performance. It is best to choose a relatively simple model. And its performance is not too bad. GBDT is selected in our experiments.

$$PRM \leftarrow Training(C_L, Y_L) \tag{1}$$

When getting trained model, the pseudo labels $Y_U$ could be gotten which can be formulated by the following:

$$Y_U = PRM(C_U) \tag{2}$$

### 3.4. The Phase II

All the data could get its corresponding label after phase I, although there may be some wrong labels. In order to alleviate the misalignment between labeled data and pseudo-labeled data, the encoder module is designed at the beginning of phase II. It can map the whole data into same space $\mathbb{R}^d$. In other words, each element $e$ in $E_L, E_U$, $e \in \mathbb{R}^d$. This can be formulated as follows:

$$E_L, E_U = Encoder(C_L, C_U) \tag{3}$$

Once the same space dataset and labels are obtained, two classifiers can perform prediction task respectively in supervised and semi-supervised ways. During our experiments, two instances of MLP classifier are adopted in our framework. $\hat{y}_L$ and $\hat{y}_L$ are the outputs of supervised classifier and semi-supervised classifier respectively. It can be shown as follows:

$$\hat{y}_L = \text{Softmax}\left(Supervised_{CLS}(E_L, Y_L)\right) \tag{4}$$

$$\hat{y}_U = \text{Softmax}\left(Semi-Supervised_{CLS}(E_U, Y_U)\right) \tag{5}$$

where $\hat{y}_L, \hat{y}_U \in \mathbb{R}^m$ denotes the probabilities of labels.

These loss functions $\mathcal{L}_L, \mathcal{L}_U$ are defined as the cross-entropy of the prediction and the ground truth, which can be written as follows:

$$\mathcal{L}_L = -\sum_{i=1}^{m} y_L^i \log(\hat{y}_L^i) + (1 - y_L^i)\log(1 - \hat{y}_L^i) + \lambda_L \|\Delta_L\|^2 \tag{6}$$

$$\mathcal{L}_U = -\sum_{i=1}^{m} y_U^i \log(\hat{y}_U^i) + (1 - y_U^i)\log(1 - \hat{y}_U^i) + \lambda_U \|\Delta_U\|^2 \tag{7}$$

where $\lambda_L$ and $\lambda_U$ are parameter specific regularization hyperparameters to prevent overfitting, while $\Delta_L$ and $\Delta_U$ are model parameters of supervised classifier and semi-supervised classifier respectively.

Although encoder module alleviates the misalignment of labeled data and pseudo-labeled data, the ASSL4CCR is still suffering from partly wrong label data. Therefore, the adversarial learning is proposed to tackle this problem. Specifically, there is an additional module named discriminator $Dis$ which aims to tell model whether current data is from labeled data or pseudo-labeled data. Hence, the adversarial loss is defined as follows:

$$\mathcal{L}_{adv} = \frac{1}{L}\sum_{e_l \in E_L}\left(\log(Dis(e_l))\right) + \frac{1}{U}\sum_{e_u \in E_U}\left(\log(1 - Dis(e_u))\right) + \lambda_{adv}\|\Delta_{adv}\|^2 \tag{8}$$

where $\lambda_{adv}$ and $\Delta_{adv}$ are regularization hyperparameter and discriminator respectively. The adversarial learning not only aligns the labeled and unlabeled data to some extent, but also tells the model the degree of trust to this answer. During the framework of ASSL4CCR, the following loss is minimized on the whole dataset:

## 4. Experiments

In this section, we introduce the extensive experiments for evaluating the performance of our proposed methods. We describe the dataset firstly, then present the experiment results of ASSL4CCR compared with baselines to prove the knowledge improvement from unlabeled data which is the main task of this work.

### 4.1. Dataset and Pre-processing

We evaluate our proposed ASSL4CCR framework on the Chinese public-listed corporate credit rating dataset which build based on the annual financial statements of corporations and the China Stock Market & Accounting Research Database (CSMRA) for the results of credit rating. Real-world data is usually noisy and incomplete. After the same data cleaning and preprocessing as work [7], we get 39 features and 9 rating labels. These features are divided into 6 categories: profit capability, operation capability, growth, capability, repayment capability, cash flow capability, dupont identity. Credit levels are from AAA, AA to CC and C. The following table and figure show details.

### 4.2. Compared with other Baselines

**Baseline Methods.** We use the following methods as baselines which achieve the state-of-the-art performance in the corporate credit rating domain:
- **LR:** Logistic regression (LR) model is a generalized linear model. It is widely used in the financial field due to its interpretability.
- **SVM:** The machine learning algorithm, Support Vector Machine with linear kernel.
- **MLP:** Multi-Layer Perceptron, A simple neural network. There are 1000 hidden units in the hidden layer and ReLU for activation function.
- **Xgboost:** eXtreme Gradient Boosting, a scalable machine learning system for tree boosting.
- **CCR-CNN:** Corporate credit model based on convolution neural networks [7].
- **CCR-GNN:** Build the graph for every corporation, then apply graph neural networks (graph attention networks) for corporate credit ratings [20].

**Metrics.** This experiment uses three common-used metrics for evaluations: precision, recall and F1-score.

To demonstrate the overall performance of our proposed ASSL4CCR framework, we compared it with mentioned baselines. The experiment results are shown as following Table 1.

Table 1. The Performance of ASSL4CCR with other Baselines

| Model | Recall | Accuracy | F1-score |
|---|---|---|---|
| LR | 0.76250 | 0.80970 | 0.81946 |
| SVM | 0.83750 | 0.89247 | 0.88961 |
| MLP | 0.91406 | 0.93568 | 0.93245 |
| Xgboost | 0.92343 | 0.94225 | 0.94133 |
| CCR-CNN | 0.92812 | 0.95253 | 0.94518 |
| CCR-GNN | 0.93437 | 0.95012 | 0.95177 |
| **ASSL4CCR** | **0.95321** | **0.96115** | **0.96252** |

As seen from the table, within the proposed ASSL4CCR framework, we establish new state-of-the-art performances of corporate credit rating, which verifies that the unlabeled data contains enormous knowledge. It can be mined to help improve the performance of credit model. This clearly demonstrates the power of the proposed ASSL4CCR which can extract useful information from enormous unlabeled data.

## 5. Conclusion

In this paper, we develop a framework, ASSL4CCR for brevity, named adversarial semi-supervised learning in terms of corporate credit rating task. The proposed ASSL4CCR effectively couples adversarial learning and semi-supervised learning to tackle this problem. This allows our model to alleviate the misalignment between labeled and unlabeled and boosts the representation and generalization of the model. What's more, our experiments prove that the proposed two-phase training is strongly beneficial for corporate credit rating by the way of mining the useful knowledge from massive unlabeled data, under the proposed ASSL4CCR framework.

## Conflict of Interest

The authors declare no conflict of interest.

## Author Contributions

Bojing Feng: Conducted the primary research, initial experiments and drafted the article.
Wenfang Xue: Provided active feedback on the drafts for revision and supported the research idea overall.
All authors had approved the final version

## Acknowledgment

This work is jointly supported by National Natural Science Foundation of China (62071017) and Major Projects in Tianjin Binhai New District (BHXQKJXM-PT-ZKZNSBY-2018001). We thank Feifei Li for her useful comments.

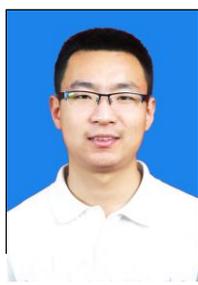

**Bojing Feng** was born in Lvliang, Shanxi, China in 1997. He has completed his bachelor's degree from Central South University, Chang Sha, China in June 2019. At present he is pursuing his master's degree in computer science in Center for Research on Intelligent Perception and Computing, National Laboratory of Pattern Recognition, Institute of Automation, Chinese Academy of Science. His research interest includes data mining, financial risk, machine learning and deep learning in finance, especially in corporate credit rating.

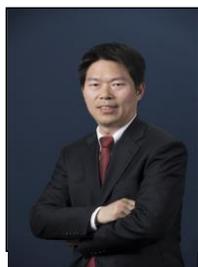

**Wenfang Xue** Dr. Xue is an associate research fellow in CRIPAC, NLPR, CASIA and an executive director in Tianjin Academy for Intelligent Recognition Technologies. Now he is integrating high-quality resources from the Chinese Academy of Sciences, the Chinese Academy of Social Sciences, Tsinghua University, Renmin University, Nankai University, China export & credit insurance corporation and other institutions to customize the national risk assessment model, industrial research model and international public opinion analysis model. And he is building a national-level intelligent risk detection, identification, prediction and early warning platform for overseas investment, and use intelligent identification technology to rapidly improve the intelligence level of Chinese think tanks and the risk management and control level of enterprises going global.